\documentclass[journal]{IEEEtran}
\usepackage{cite}

\ifCLASSINFOpdf
\else
\fi
\usepackage{amsmath}
\usepackage{algorithm} 
\usepackage{algpseudocode} 

\usepackage{url}

\usepackage{array}

\ifCLASSOPTIONcompsoc
 \usepackage[caption=false,font=normalsize,labelfont=sf,textfont=sf]{subfig}
\else
 \usepackage[caption=false,font=footnotesize]{subfig}
\fi

\usepackage{stfloats}
\usepackage{url}

\usepackage{graphicx}

\begin{document}
%
\title{Scalable Nanophotonic-Electronic Spiking Neural Networks}
%
%
%

\author{Luis El Srouji,
        Yun-Jhu~Lee,
        Mehmet Berkay~On,
        Li Zhang,
        and~S.J.~Ben~Yoo,~\IEEEmembership{Fellow,~IEEE,~Fellow,~Optica}}%

\IEEEspecialpapernotice{(Invited Paper)}

\maketitle

\begin{abstract}
Spiking neural networks (SNN) provide a new computational paradigm capable of highly parallelized, real-time processing. Photonic devices are ideal for the design of high-bandwidth, parallel architectures matching the SNN computational paradigm. Co-integration of CMOS and photonic elements allow low-loss photonic devices to be combined with analog electronics for greater flexibility of nonlinear computational elements. As such, we designed and simulated an optoelectronic spiking neuron circuit on a monolithic silicon photonics (SiPh) process that replicates useful spiking behaviors beyond the leaky integrate-and-fire (LIF). Additionally, we explored two learning algorithms with the potential for on-chip learning using Mach-Zehnder Interferometric (MZI) meshes as synaptic interconnects. A variation of Random Backpropagation (RPB) was experimentally demonstrated on-chip and matched the performance of a standard linear regression on a simple classification task. Meanwhile, the Contrastive Hebbian Learning (CHL) rule was applied to a simulated neural network composed of MZI meshes for a random input-output mapping task. The CHL-trained MZI network performed better than random guessing but does not match the performance of the ideal neural network (without the constraints imposed by the MZI meshes). Through these efforts, we demonstrate that co-integrated CMOS and SiPh technologies are well-suited to the design of scalable SNN computing architectures.
\end{abstract}


%
\IEEEpeerreviewmaketitle

\section{Introduction}

Computation using spiking neural networks (SNN) yields three major architectural advantages: (1) the sparsity of communication between elements which reduces energy cost, (2) the binarization of communication without discretization of messages (i.e. all-or-nothing spike responses), and (3) completely asynchronous operation of computational units. At the architectural level, the spiking paradigm requires several computational elements in common to the traditional artificial neural network (ANN)---weighted addition, nonlinearity, and learning algorithms---though with the additional complexity of computation spread through time. Traditional computational approaches based on the von Neumann computing architecture---including modern system architectures equipped with graphical processing units (GPUs)---are not well-suited for the use of this computational paradigm due to the fundamental separation between computing and memory units and resulting serialization of many processing tasks. In turn, the traditional computing paradigm cannot efficiently support the requisite computational elements without significant simplification or long latencies thus warranting the development of new computer architectures. Neuromorphic design operates under the general principle that evolution has already produced a successful SNN architecture for operating under real-time, low-power conditions. Approaches to replicating this design employ a variety of digital, analog, or mixed-signal circuits that can be based on electronic, photonic, or optoelectronic devices. Nonetheless, substantially more work is necessary to determine the optimal approach to abstract, apply, and improve upon this evolutionary design. 

Digital neuromorphic processors (such as TrueNorth\cite{Merolla2014}, Loihi\cite{Davies2018Loihi:Learning}, SpiNNaker\cite{Painkras2012SpiNNaker:Simulation}, etc.) increase the parallelization of processing by including a large number of cores that allow asynchronous computation---in contrast to GPU architectures---though this approach is not unlike a specialized and monolithic form of cluster computing. Though each core completes its operation in parallel, a desire for determinism in digital electronics necessitates synchronization between simulated time steps. This, in turn, limits full asynchronous operation which may prove to be prohibitive at biological network scales. On the other hand, analog electronic meshes can provide fully parallel computation, though the capacitance of electrical wire networks causes increases to both latency and power consumption.

Photonic and optical computing efforts have sought to exploit the nearly lossless and parallel communication capabilities of optical fibers into the domain of photonic integrated circuits (PICs). A number of demonstrations have already shown matrix multiplication and convolutional processing using non-spiking photonic circuits \cite{Mehrabian2018PCNNA:Accelerator,Xu202111Networks,Feldmann2021ParallelCore}. These devices use a combination of wavelength-division multiplexing (WDM) and space-division multiplexing (SDM) to manage multiply-and-accumulate (MAC) operations in parallel; thus, these schemes are also compatible with spike processing in synaptic networks. Choices of nonlinearity in spiking elements varies widely from one approach to another, though a major division can be made between all-optical and optoelectronic approaches. Optical nonlinearities typically have shorter lifetimes and can potentially service higher speed computation compared to electronic nonlinearities based on electronic charges or currents. However, the manipulation of these nonlinearities is governed mainly by material properties which are fixed after fabrication. Given that biological neural networks operate over a range of time-scales, it is preferable to have programmable elements in the neuron design. Optoelectronic approaches can take advantage of recent progress in the co-integration of CMOS circuitry with photonic devices to form flexible and programmable spiking neuromorphic computers.

In addition to the architectural benefits, SNNs offer provable advantages in solving graph algorithms, constraint satisfaction, and other optimization problems \cite{Chou2018OnNetworks,Verzi2018ComputingTiming,Kwisthout2020OnNetworks,Aimone2021ProvableNetworks}. Incorporating learning and training using Hebbian\cite{Gerstner2002MathematicalLearning} and spike-timing-dependent plasticity (STDP)\cite{Caporale2008SpikeRule} algorithms also allows for the application of SNNs in many of the same contexts as deep neural networks (DNN). These learning rules have the additional architectural advantage of using only locally available information for the updating of each synapse. In principle, this means that all weight updates within the network can be calculated completely in parallel. With the appropriate network topology and training signals, Hebbian learning has also been shown capable of error-driven learning equivalent to backpropagation in deep and convolutional neural networks of moderate size \cite{OReilly1996,Amato2019HebbianNetworks}.

In this paper we will discuss the design of a nanophotonic-electronic neuromorphic architecture for native SNN computation with on-chip learning. Sec.~\ref{Background} will provide a brief taxonomy of existing photonic and optoelectronic approaches to spiking neuron and optical matrix multiplication. Next, Sec.~\ref{Technologies} will discuss the technologies and algorithms used, while addressing scalability and remaining design challenges. Finally, Sec.~\ref{Perspectives} will detail future directions and perspectives for the design of photonic neuromorphic processors.

\section{Background and Survey} \label{Background}

Spiking neural networks require two primary computational elements: (i) a nonlinear spiking unit that can integrate its inputs over time (the neuron) and (ii) a reconfigurable network to service weighted connections between these elements (the synaptic network). As previously alluded, the nonlinearities exploited for the design of spiking units can vary between all-optical and optoelectronic approaches, the choice of which can limit the choice of network elements to service communications between units. 


    \subsection{Spiking Nonlinearity}

Excitability describes the ability of a system to quickly and temporarily deviate from its quiescent state following small perturbations and can be rigorously described through bifurcation analysis as done by Izhikevich \cite{IZHIKEVICH2000NEURALBURSTING}. Biological neurons are dynamical systems and have been classified into saddle-node and Andronov-Hopf bifurcations which correspond to \emph{integrator} and \emph{resonator} neurons respectively. Simply put, integrator neurons integrate their inputs and will generate a spike upon reaching some dynamic threshold, while a resonator neuron undergoes some internal subthreshold oscillation with an increased response and likelihood to generate a spike for inputs that fall at specific phases of a resonant frequency.

Computationally useful spiking neurons, however, need not be entirely biologically plausible. Instead, behavior is commonly summarized by the \emph{leaky-integrate-and-fire} (LIF) neuron model. In the LIF model, the membrane potential constantly undergoes exponential decay towards its \emph{resting potential} with discrete jumps at each input spike. When the membrane potential reaches a fixed threshold, the spike is generated and the potential is instantaneously returned to a \emph{reset potential}. LIF neurons are only able to represent integrator neurons and lose much of the complexity of behaviors seen in biological neurons. Alternatively, Izhikevich devised a neuron model which faithfully reproduces a wide range of biologically observed behaviors using only four parameters and two coupled differential equations \cite{Izhikevich2003SimpleNeurons}. For a brief summary of computationally relevant neuron behaviors and a comparison of neuron models see \cite{Izhikevich2004WhichNeurons}. Other taxonomies exist to classify neuron types according to these behaviors, though some evidence has shown that biological neurons may flexibly switch between these types based on the history of the cell \cite{Steriade2004NeocorticalEntities}. As such, an ideal hardware implementation of spiking neurons would be capable of representing a range of neuron types for maximal computational ability.

A number of semiconductor lasers have been explored which create isomorphisms between the time dynamics of material parameters of active photonic elements and the cellular mechanisms of biological neurons. Researchers have exploited the time dynamics of photocarriers, thermal diffusion, optical modes, and polarization competition to create excitable laser devices with varying degrees of faithfulness to the biology. Photonic spiking neurons can be most meaningfully divided into two categories based on whether the device can accept optical or electrical inputs---some devices can be modulated by either, but electrical input may be preferred for the advantages in system design discussed in Sec.~\ref{Networks}. 

Optical devices can be further classified into coherent and incoherent devices based on how incoming wavelengths are used to excite the active medium. In coherent excitable semiconductor lasers \cite{Giudici1997AndronovFeedback,Coomans2011SolitaryNeurons,Brunstein2012ExcitabilityNanocavity,Dambre2013ExcitabilityResponse,Garbin2014IncoherentLaser}, the incoming signal interacts with a lasing cavity mode on the same wavelength to modulate the output signal directly. Excitability is induced by disturbing the balance between competing modes or polarizations which, with sufficient input energy, temporarily drive the extinction of one mode and amplification of the other. Bandwidth for such devices is bound by the cavity Q factor, with a time constant for energy dissipation given by $\tau = Q/\omega_0$. For incoherent devices \cite{Nahmias2013AComputing,Selmi2014RelativeLaser,Hurtado2015ControllableSystems,Selmi2015TemporalLaser} the incoming signal interacts with some element within the cavity that indirectly modulates the output signal. This may take the form of optical pumping of the laser medium, or otherwise modulating the carrier populations which affect gain and saturation properties. Bandwidth for such approaches are limited by the dynamics of these carrier populations which are material dependent. Alternatively, optoelectronic approaches \cite{Romeira2013ExcitabilityPhoto-detectors,Tait2015ExcitableSimulation} can allow for the design of analog circuitry with time-dynamics that can be fit to a variety of available neuron models, with lasers modulated by current injection in response to processed photodetector input. Optoelectronic designs are mainly limited by the total bandwidth of integrated photodetectors and electronics, though some estimates suggest that bandwidths upwards of 10 GHz can be expected; see \cite{Tait2016RecentProcessing} for a more in-depth review of various excitable semiconductor lasers with discussions of bifurcation paralleling Izhikevich's analysis.
        
    \subsection{Reconfigurable Networks} \label{Networks}
    
Given the ability of silicon waveguides to simultaneously support a wide range of wavelengths with negligible loss, on-chip optical networks are most efficiently parallelized using wavelength division multiplexing (WDM). Time-division multiplexing (TDM) offers another scheme for sharing computing resources over time, but the asynchronous and stochastic nature of SNNs is not likely to benefit from this technique. Using WDM, signals from each neuron can be routed according to wavelength and resources for matrix multiplication may potentially be used for multiple independent operations to support weight sharing and convolution. To support such architectures, different neurons must be distinguishable by output wavelength. However, the system does not need a unique wavelength for each neuron since most SNN architectures group neurons into layers that provide an additional level of hierarchy for routing structures. 

Using a WDM approach, arrayed waveguide grating routers (AWGR) can be used to support all-to-all routing schemes between neural layers \cite{Mitsolidou2018SiliconCommunication,Zhang2019Foundry-EnabledTransceivers,Xiao2019Flex-LIONS:Fabric}. Inputs to each layer would be passed through reconfigurable optical matrix multipliers such as cross-bar networks, micro-ring resonator (MRR) banks, and mach-zehnder interferometry (MZI) meshes. MZI meshes can perform unitary matrix transformations that correspond to lossless multiplication and are thus particularly suitable for low-power neuromorphic computing. See \cite{Srouji2022PhotonicComputing} for a longer discussion on the design trade-offs between each of these devices. Sec.~\ref{Synapses} describes our MZI mesh architecture, while Sec.\ref{Training} details algorithms for training SNNs using MZI meshes.

\section{Scalable Photonic SNN Technologies} \label{Technologies}

    \subsection{Towards Attojoule nanophotonic-electronic spiking neurons}

Neurons provide nonlinearity and signal regeneration between each neural network layer. Our previous work \cite{Lee2022PhotonicModel} presents an optoelectronic neuron design with projected energy efficiency on the order of $200\,aJ/\text{spike}$. Because the time-scales of electrical circuits are more tunable than photonic nonlinear materials, the neuron is more easily programmable while still taking advantage of low-loss communication provided by photonic interconnects. This design closely matches the behavioral characteristics of the Izhikevich neuron model to achieve a variety of neural behaviors. To move a step forward in realizing attojoule energy efficiencies, we have updated this design on a more advanced foundry platform.  

        \begin{figure}[htp]
        \centering
        \includegraphics[width=8cm]{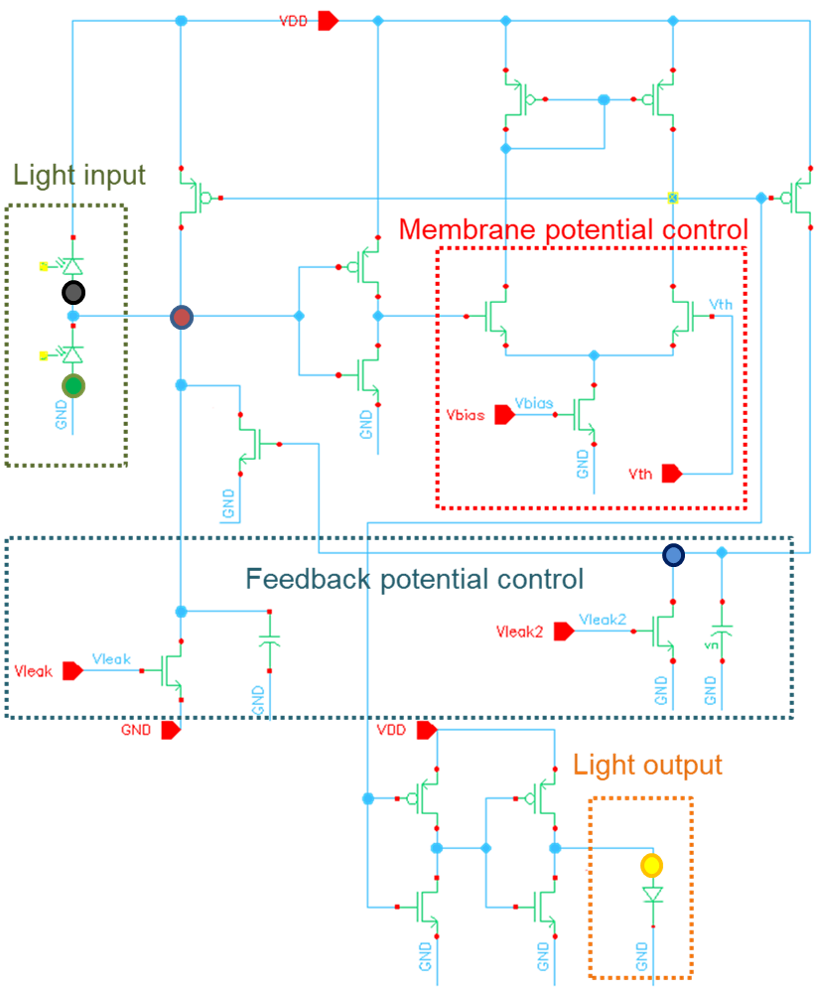}
        \caption{The circuit diagram of 45SPCLO neuron design. The circuit mechanism of optoelectronic neuron start with converting light input to current. The membrane potential control section will decide the neuron threshold, feedback strength to refractory feedback potential control section and send the light out from laser diode. The feedback potential control decides the refractory strength and the frequency of spiking.}
        \label{fig:45clo_neuron_circuit}
        \end{figure}
        
Our previous foundry neuron design \cite{Lee2022PhotonicModel} also employs optoelectronics and a scalable MZI interconnect mesh, however, this design is not capable of the full range of neural behaviors described by the Izhikevich model. Using the GlobalFoundries (GF) 45SPCLO PDK, a new neuron was designed that can support a wider range of neural behaviors depending on applied voltage biasing. GF 45SPCLO is the successor of the GF 90WG PDK, and preserves the same CMOS-silicon photonic co-integration with a more advanced process node and additional metal routing layers. Fig.~\ref{fig:45clo_neuron_circuit} shows the GF 45SPCLO neuron circuit design. The labeled red pins mark  voltage biasing nodes that can be adjusted to achieve the desired neuron behavior. These nodes correspond to the control of an adjustable positive bias ($V_{bias}$), spiking threshold ($V_{th}$), refractory feedback rate ($V_{leak}$), and adaptation rate ($V_{leak2}$). The function of these node voltages is divided between membrane potential control and feedback potential control. Membrane potential controls $V_{bias}\;\&\;V_{th}$ adjust the spiking threshold and determine the current flow into membrane potential for each spike input. Feedback potential controls, $V_{leak}\;\&\;V_{leak2}$, determine the strength of negative feedback on the membrane potential and the length of refractory period. Balanced photodetectors receive excitatory and inhibitory light input. The diode at the circuit output incorporates the I-V characteristics of the laser diode chosen for the design.

\begin{figure}[!htp]
        \centering
        \includegraphics[width=8cm]{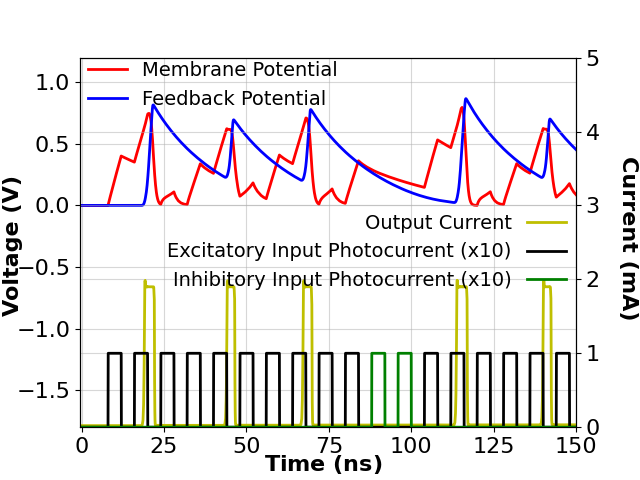}
        \caption{Basic spiking behavior with excitatory and inhibitory input. inhibitory inputs are assigned on \#11 and \#12 spike which oppose the excitatory input currents.}
        \label{fig:basic_inh}
        \end{figure}

 To demonstrate this design, we first simulate the basic spiking behavior in response to excitatory and inhibitory inputs simulated in Cadence Spectre (shown in Figure \ref{fig:basic_inh}). The nodes of each measurement are matched to the color of each line in Figure \ref{fig:45clo_neuron_circuit}. We include inhibitory inputs on spike \#11 and \#12 and can confirm from Fig.~\ref{fig:basic_inh} that inhibitory input suppressed the membrane potential and output, which matches our expectation.
 
\begin{figure}[!htp]
        \centering
        \includegraphics[width=8cm]{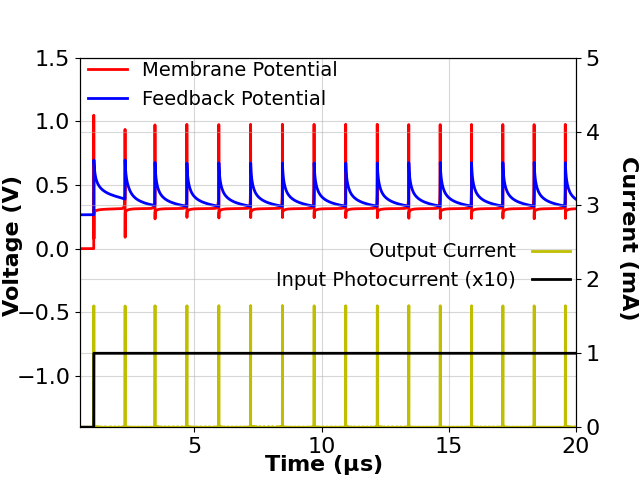}%
        \caption{Regular spiking neuron behavior. The step input shows that the circuit feedback mechanism properly functions and that the neuron is an excitable system. The spiking rate for regular spiking is set to the lower-end of each voltage supply.}
        \label{fig:RS}
        \end{figure}

\begin{figure}[!htp]
        \centering
        \includegraphics[width=8cm]{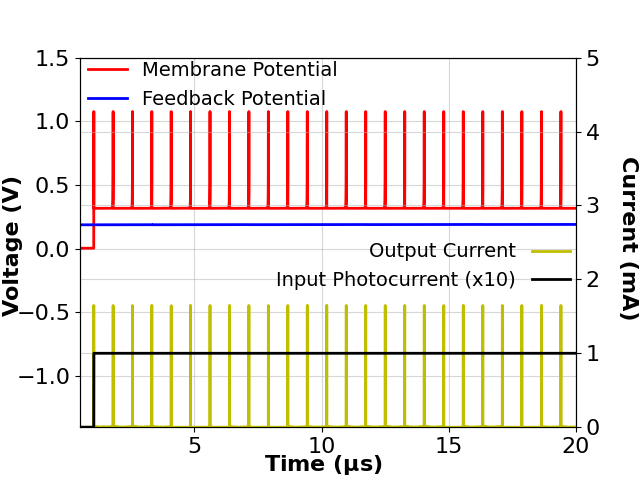}%
        \caption{Fast spiking neuron behavior. Each spike is 50ns faster than regular spike. The spiking speed can be adjusted by changing the voltage supplies. $V_{th}$ has most influence on spiking rate adjustment.}
        \label{fig:FS}
        \end{figure}

\begin{figure}[!htp]
        \centering
        \includegraphics[width=8cm]{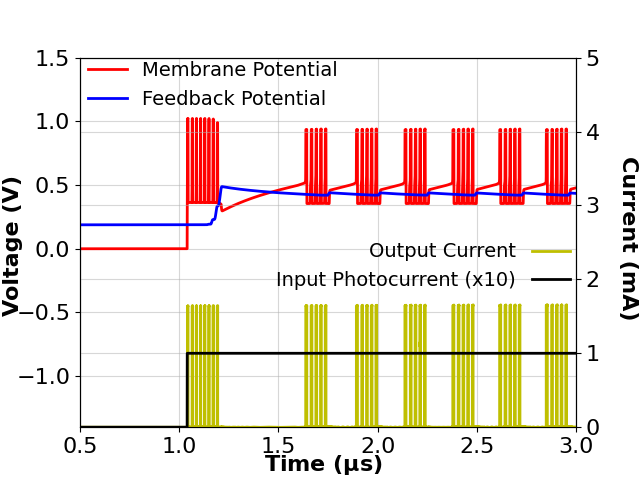}%
        \caption{Chattering neuron behavior. The neuron continuous firing for $0.3\,\mu s$ and resting $0.6\,\mu s$. Then repeat this cycle with shorter firing and resting period.}
        \label{fig:chattering}
        \end{figure}

Next, we demonstrate three spiking patterns: regular spiking (RS), fast spiking (FS), and chattering (CH) in analogy to \cite{Izhikevich2003SimpleNeurons}. These behaviors can be achieved flexibly by modifying the voltages at each biasing pin, which allows a greater tolerance for mismatch between design and tapeout. These spiking patterns are shown in Figure \ref{fig:RS}, Figure \ref{fig:FS}, and Figure \ref{fig:chattering} respectively. Input photocurrents are simulated as step functions from $0.0\,mA$ to $0.1\,mA$, node voltages corresponding to each behavior are set as follows:

1)	Regular spiking: bias ($V_{bias}$) low, threshold ($V_{th}$) low, refractory feedback ($V_{leak}$) low, and frequency adaptation ($V_{leak2}$) low.

2)	Fast spiking: bias ($V_{bias}$) low, threshold ($V_{th}$) high, refractory feedback ($V_{leak}$) low, and frequency adaptation ($V_{leak2}$) high.

3)	Chattering: bias ($V_{bias}$) medium, threshold ($V_{th}$) medium, refractory feedback ($V_{leak}$) high, and frequency adaptation ($V_{leak2}$) high.

These simulations verify the ability of the neuron circuit to achieve various spiking patterns on the more advanced 45SPCLO process.

                
    \subsection{Photonic MZI Mesh as Synaptic Network} \label{Synapses}

\begin{figure}[ht]
    \centering
    \includegraphics[width=\linewidth]{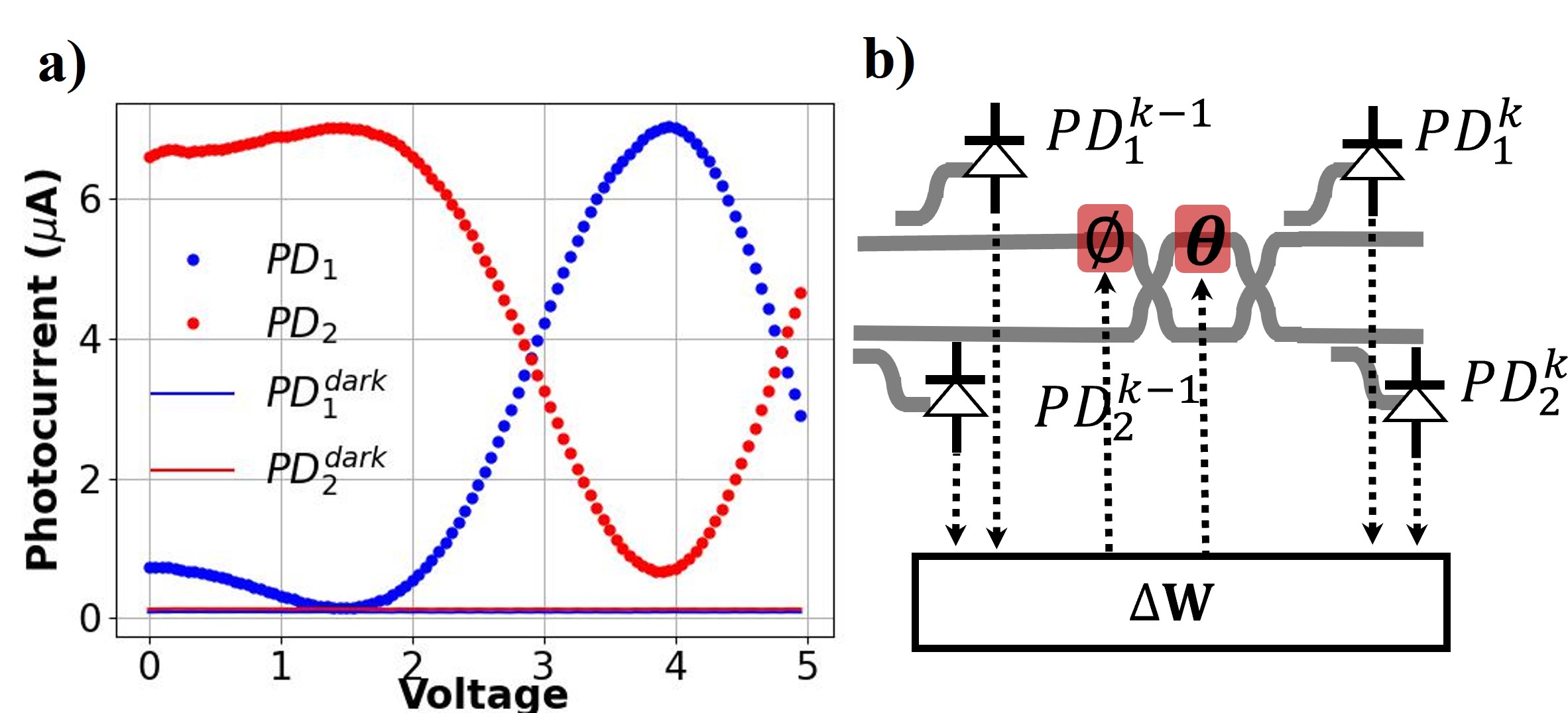}
    \caption{a) DC voltage sweep for phase shifter $\theta$, b) 2$\times$2 MZI unit with power monitoring and local training features}
    \label{fig:mzi}
    \end{figure}

    The building block of an MZI Mesh is a 4-port device that consists of two 50:50 beam splitters and two-phase shifters, $\theta$ and $\phi$ as shown in Fig.~\ref{fig:mzi}~(b). The phase shifter $\theta$, inside the interferometer, controls the power splitting ratio. Meanwhile, the phase shifter, $\phi$, outside of the interferometer, controls the relative phase difference between the two coherent input ports. As demonstrated in Fig \ref{fig:mzi}~(a), the tunable power splitting functionality is tested by sweeping applied DC voltage on the phase shifter $\theta$. MZI Meshes can be arranged in several ways, with the most popular arrangements being the triangular\cite{Reck1994ExperimentalOperator} or rectangular\cite{Clements2016OptimalInterferometers} formations. Both of the formations can realize an arbitrary N$\times$N unitary matrix. There are a variety of applications where MZI Meshes are employed, such as mode-division multiplexing \cite{Choutagunta2020AdaptingLinks}, free-space beamforming\cite{Milanizadeh2021MultibeamMesh}, quantum computing \cite{Qiang2018Large-scaleProcessing}, and photonic neural networks \cite{Shen2017DeepCircuits}. Our work utilizes MZI Meshes as synaptic interconnections for bio-inspired neural networks and aims to integrate learning algorithms on the same chip. Although calibration procedures of the MZI Meshes are well-studied \cite{Pai2019ParallelNetwork}, training MZI Meshes as neural network (NN) interconnects remains challenging. Hughes \textit{et al.} \cite{Hughes2018TrainingMeasurement} proposed an in-situ training to realize the traditional backpropagation algorithm for MZI meshes, and recently Pai \textit{et al.} \cite{Pai2022ExperimentallyNetworks} experimentally demonstrated the method. This in-situ training requires additional forward and backward light propagation with power monitoring for each phase shifter element at each step. There are various approaches to monitor power. For example, Pai \textit{et al.} \cite{Pai2022ExperimentallyNetworks} utilized power tapping and grating couplers with an infrared camera to record the emitted power from MZI Meshes. Alternatively, Morichetti \textit{et al.} \cite{Morichetti2014Non-invasiveMonitoring} used a non-invasive power sensing device introduced for silicon waveguides. We exploited 1:99 power taps and Ge photodetectors (PDs) which are  available as an instance in Process Design Kit (PDK) elements of the active silicon photonic multi-project-wafer (MPW) runs from the AIM Photonic foundry. Fig.~\ref{fig:mzi}~(a) shows photocurrent changes on the monitoring PDs with respect to applied voltage on phase-shifter $\theta$. Although we used thermo-optics as a simple and practical phase-shifting mechanism, it is possible to utilize micro-electro-mechanical systems (MEMS) for even lower power consumption \cite{Yuji2020MORPHIC:MEMS} in future designs.
        
\begin{figure*}[ht]
    \centering
    \includegraphics[width=\linewidth]{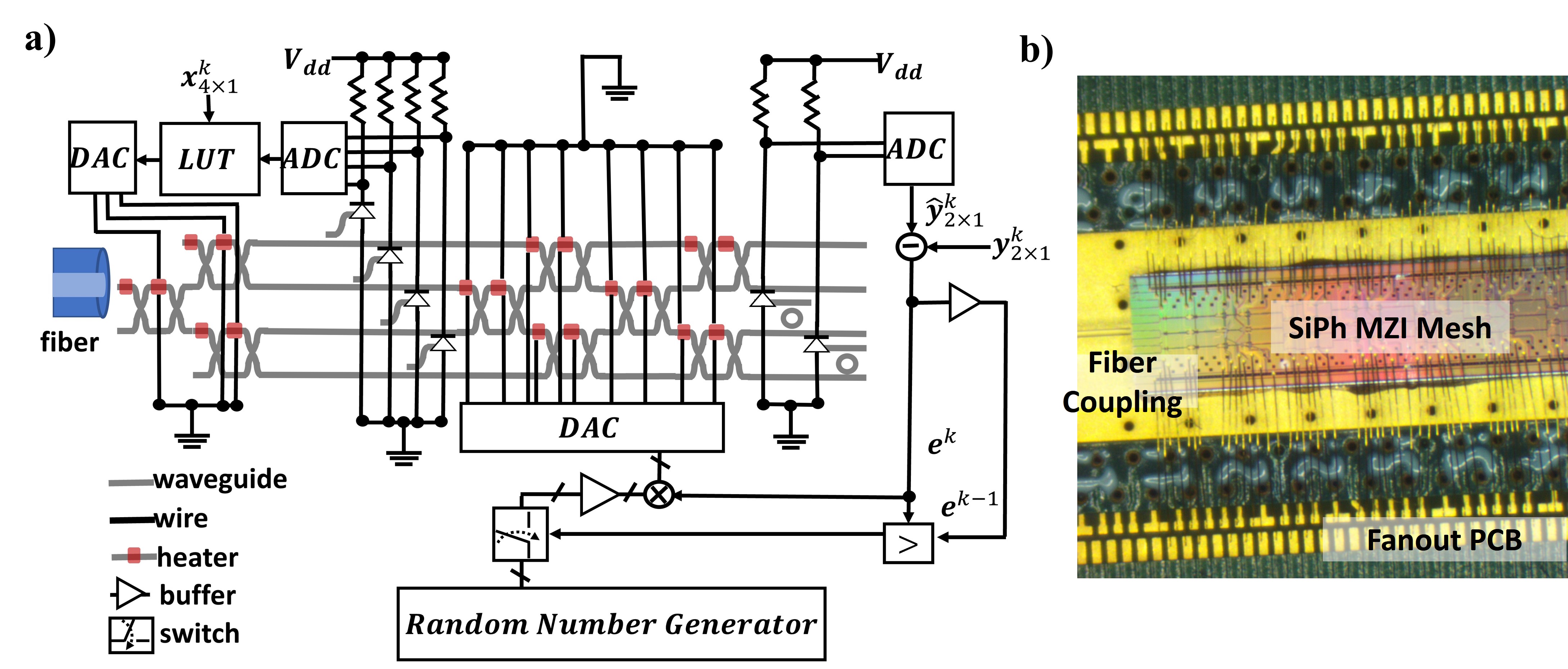}
    \caption{a) Random backpropagation experimental setup with optical and electrical components, b) SiPh MZI Mesh, wirebonded on fanout PCB}
    \label{fig:randombackprop}
    \end{figure*}

    Fig.~\ref{fig:randombackprop}~(b) shows the fabricated and tested 6$\times$6 rectangular MZI Mesh with power taps after each 2$\times$2 MZI unit as shown in Fig.~\ref{fig:mzi}~(b). At each output waveguide of the 6$\times$6 mesh, a micro ring resonator (MRR) add-drop filter is placed with a PD on the drop ports, allowing for output monitoring by either optical or electrical means. When the MRR is at resonance, the output can be monitored and accessed through the electrical interface during the training. Alternatively, the MRR resonance wavelength can be tuned to let the optical signal propagate after the MZI Mesh. In this way, multiple MZI Mesh layers can be cascaded for DNN-like implementation. All the components are available in AIM Photonic's PDK v4.0. The device is wirebonded on a fanout printed-circuit board. A USB interfaced multi-channel high current output digital to analog converter (DAC) unit drives the thermo-optic heaters and MRR add-drop filters. Similarly, the photocurrents are digitized by a USB interfaced multi-channel 250kSps analog-to-digital converter (ADC), as shown in Fig.~\ref{fig:randombackprop}~(a).
    
    \subsection{Training and Inference} \label{Training}
    
    For the on-chip training demonstration, we targeted a linear classification problem with 4-dimensional input vectors and two output classes. We used the Iris flower dataset \cite{FISHER1936THEPROBLEMS}, consisting of 3 classes and 150 input samples. For simplicity in the proof-of-principle demonstration, we excluded one of the classes that is linearly separable from the other two classes. Therefore, a linear regression classifier can achieve a maximum of 94 true classifications over 100 samples. 
    We use a single-mode cleaved fiber to couple a CW tunable laser source operating at $1553.7\,nm$ to the chip. After the edge coupler, the first three 2$\times$2 MZI stages act as tunable beam splitters and were used to generate coherent input vectors. First, the input generator phase-shifters are optimized adaptively to create desired 100 input samples. Next, optimum voltage values are recorded in a look-up table (LUT) to recall in the training and interference cycles.
    
    One of the challenges of using MZI Meshes as a synaptic weight matrix is that controllable variables (phase shifters) do not explicitly map to individual weight matrix entries. In other words, adjusting a single phase shifter will affect multiple weight matrix entries. Clements \textit{et al.} \cite{Clements2016OptimalInterferometers} devised a decomposition method for rectangular meshes. In machine learning, however, the optimum weight matrix is unknown at the beginning of training and the additional resources for continual adjustment and decomposition become intractable. Hughes \textit{et al.} \cite{Hughes2018TrainingMeasurement} demonstrated a method of differentiating the weight matrix w.r.t. each phase shifter. However, this method requires two optical propagation steps in addition to the initial inference step: one forward, and one backward. Therefore, an external controller is required to schedule each propagation, and light sources must be bidirectional. Moreover, during the additional optical propagation steps, power must be monitored for every phase shifter element. The number of phase shifters in the MZI mesh scales as $N(N-1)$ for N$\times$N weight matrices, meaning $\mathcal{O}$($N^2$) power monitoring is required. This presents remaining challenges for scalability in deep neural networks.
    
    Here, we looked for more hardware-friendly solutions and, taking inspiration from biology, explored \textit{random backpropagation} (RBP) and \textit{contrastive Hebbian learning} (CHL) for MZI Meshes. In Section \ref{RPB} we present an experimental demonstration of random backpropagation training for a linear classification task; Section \ref{Local} discusses the CHL algorithm and its relevance to human-like predictive error-driven learning. 
    
    \subsubsection{Random Backpropagation}\label{RPB}
    
    In RBP, global error is backpropagated electrically from the end of the network. As such, RBP does not require optical backpropagation or power monitoring for each individual 2$\times$2 MZI unit. An important difference between conventional BP and RBP is the direction of the gradient. BP follows the steepest gradient direction, which requires error to multiply the conjugate transpose of the forward weight matrix. In a digital computer, these forward weights are available in the memory unit, but for MZI Meshes, optical light would be physically backpropagated as discussed earlier. The original researchers demonstrated that a random backward weight matrix could also guarantee learning unless random backward weights are exactly orthogonal to the steepest backward weights \cite{Lillicrap2014RandomNetworks}. Further, neuroscience studies observed that backward synaptic connections of neural networks in mammals are not fully symmetric \cite{VanSchaik2017Event-DrivenMachines, Detorakis2019ContrastiveWeights} giving biological credibility to the RBP algorithm. \textit{Direct feedback alignment}, a variant of RBP, has also been demonstrated for MRR-based photonic weight matrices \cite{TrondheimDirectNetworks}. Given that tunable elements in the MRR bank have a one-to-one mapping with the synaptic weight matrix, it is computationally easier to calculate steepest gradient direction. Therefore, RBP can be more useful for MZI mesh training where this mapping is non-trivial. Nonetheless, MZI meshes are preferred for their ability to perform lossless matrix multiplication.

\begin{figure}[!h]
    \centering
    \includegraphics[width=9cm]{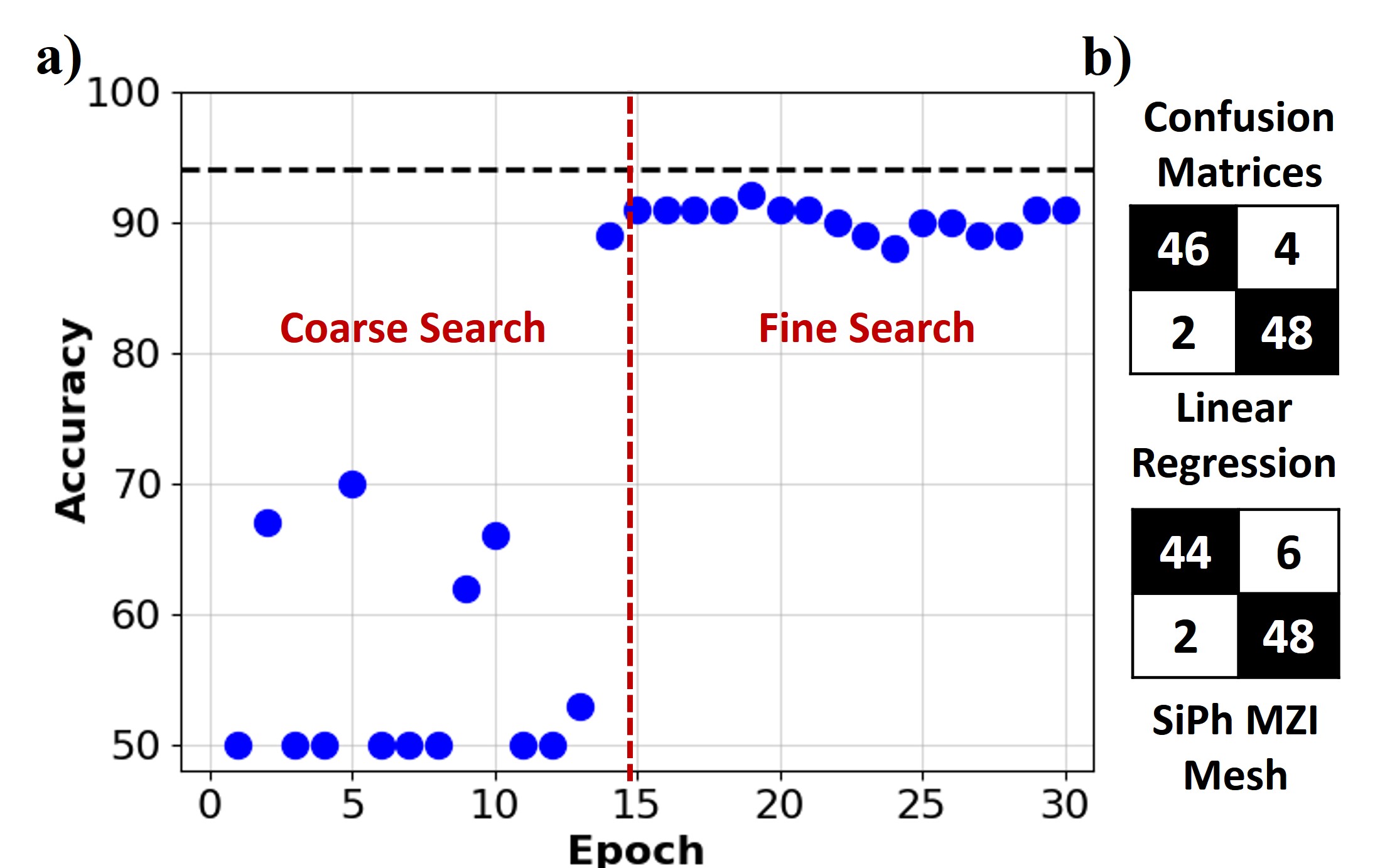}
    \caption{a) Interference accuracy during the random backpropagation training, b) Confusion matrices for ideal linear regression and SiPh MZI Mesh classifer}
    \label{fig:acc_mzi}
    \end{figure}

     Appendix \ref{RBP_Pseudocode} summarizes our method of applying RBP on a SiPh MZI Mesh, while an illustration of our experimental setup is shown in Fig.~\ref{fig:randombackprop}~(a). The multiplication, addition, comparator, and memory buffer operations are realized in an external computer through Python scripts.  Unlike conventional RBP, we draw a new random backward matrix for each iteration where the error is larger than the previous. With this modification, we empirically observed faster convergence to the classifier's highest accuracy and the ability escape local minimums, as seen in the course search of Fig.~\ref{fig:acc_mzi}~(a). Note, however, this additional operation may not be necessary for a network with a larger number of parameters and multiple synaptic layers. For example, in the papers \cite{Lillicrap2014RandomNetworks, VanSchaik2017Event-DrivenMachines, Detorakis2019ContrastiveWeights, TrondheimDirectNetworks} the authors use fixed random backward weights. Future efforts will involve the real-time implementation of these operations by integrated electronic circuits within the mesh.
    
    Fig.~\ref{fig:acc_mzi}~(a) shows the interference accuracy of the SiPh MZI Mesh classifier for each epoch. In each epoch, 100 samples are forward propagated once. We use i.i.d. random backward weights uniformly distributed in the interval $[-\mu,\mu]$. During the \textit{coarse search} cycle ($\mu=0.05$), the classifier searches different local minimums, and after some epochs, the interference accuracy decreases due to large variance on random weights. We defined an accuracy limit (85 true labels among 100 samples) and switched to the \textit{fine search} cycle ($\mu=0.0025$) when the limit was reached. As seen in Fig.~\ref{fig:acc_mzi}~(a), the coarse search cycle ended when the classifier labeled 89 samples correctly, and in the fine search cycle, 92 true labels were achieved. The confusion matrix for the ideal linear regression classifier and SiPh MZI Mesh classifier are presented in Fig.~\ref{fig:acc_mzi}~(b). The SiPh MZI Mesh misclassified only two samples compared to the ordinary least squares linear regression model we built in the computer via \textit{scikit-learn} Python package. We also implemented a numerical simulation for the MZI Meshes on the computer. From the simulation results, we observed that the SiPh MZI Meshes achieve the same accuracy as the linear regression model. Therefore, we concluded that reason for the misclassification of two input samples related to hardware imprecisions such as noise on the output PDs, electrical wires, thermal crosstalk between the phase shifters, etc.
    
    Intuitively, traditional BP outperforms RBP in terms of convergence speed due to the steepest gradient direction. However, RBP is more hardware-friendly given that forward weights are unavailable and since phase shifter-to-weight mapping is not explicit in the MZI Meshes. Because the steepest direction for the gradient is not calculated, RBP does not require any power monitoring inside the MZI Meshes except for the input and output stages. Therefore, the PDs can scale with \textit{$\mathcal{O}$($N$)} for N$\times$N weight matrices. In the future, we plan to study RBP for larger SiPh MZI Meshes and more complex machine learning problems.

    \subsubsection{Contrastive Hebbian Learning}\label{Local}
    
In contrast to backpropagation where learning is based on credit towards global error, learning in biological systems is restricted to information local to a given synapse. Despite this, biological neural networks are able to autonomously develop expansive hierarchical abstractions of information useful for interpreting the environment. This represents a form of self-supervised learning that needs no explicit calculation of error, but instead relies on chemical signals marking recent spiking activity local to a synapse.

\begin{figure}[!h]
\centering
\includegraphics[width=\linewidth]{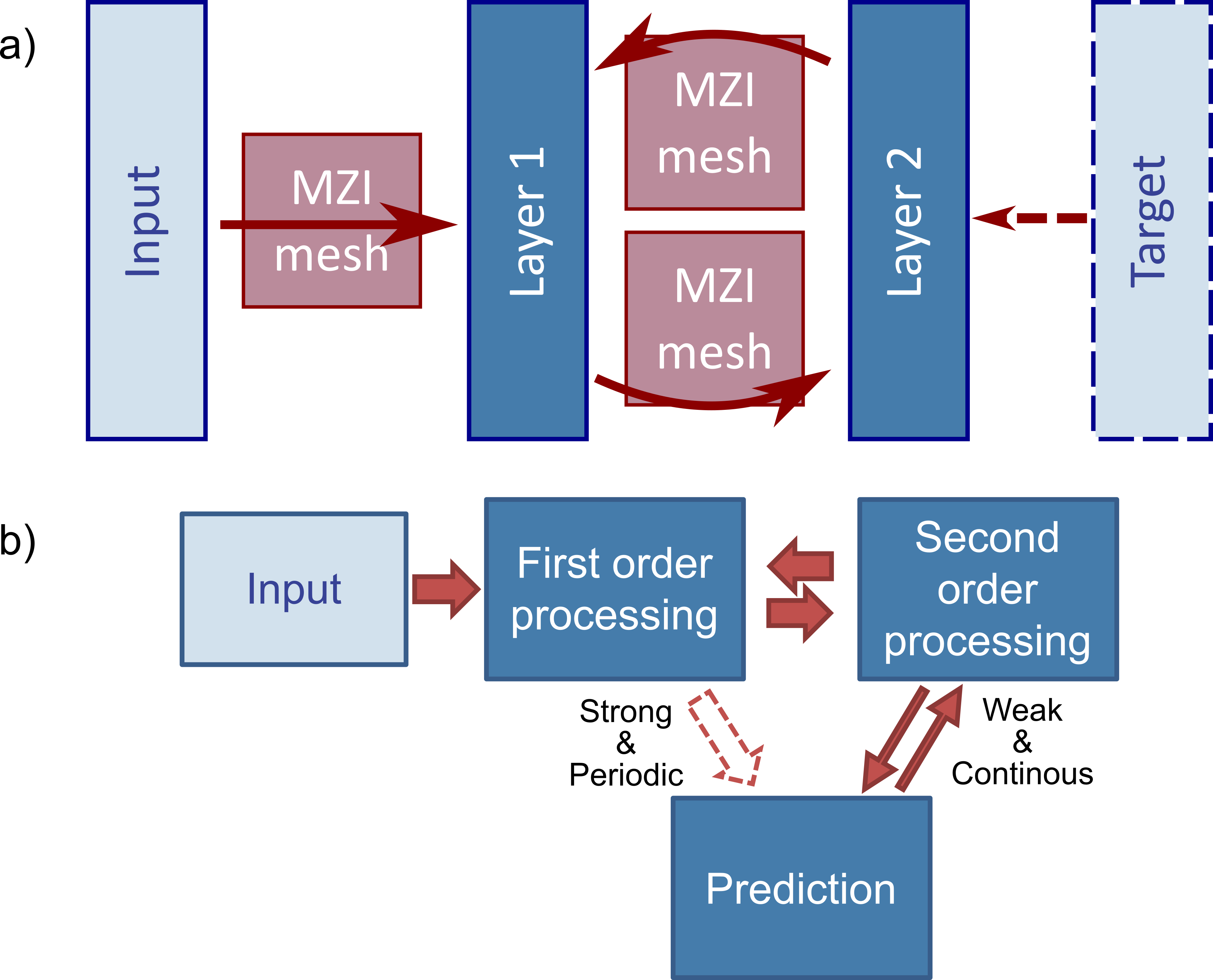}

\caption{a) Schematic of two-layer CHL network structure. b) Extension of this structure to predictive error-driven learning.}
\label{fig:LocalLearning}
\end{figure}

O'Reilly \cite{OReilly1996} proved that differences in activity at two distinct phases of network computation can drive a class of temporal-difference learning rules that is equivalent to backpropagation and gradient descent of errors. This equivalence, however, only holds for a multi-layer perceptron (MLP) with recurrent feedback connections between each layer as in Fig.~\ref{fig:LocalLearning}~(a). The general learning rule has minor variations which have different properties, though an empirical test under common MLP tasks showed that the CHL variant often converges to a solution most quickly:
\begin{equation} \label{eq:CHL}
    \Delta w_{ij} = \eta(a_i^+a_j^+ - a_i^-a_j^-)
\end{equation}
where $a_i$ and $a_j$ are variables representing the activity of the $i$th and $j$th neuron, and $\eta$ dictates the rate of learning. 

Superscripts denote the phase of activity that each variable represents. The minus phase of execution occurs first, and represents the network's natural response to the given input sample. Next, in the plus phase, the target activity is imposed on the output layer and the network reaches a new equilibrium. For fastest implementation, the duration of each phase should be the minimum time required for stable output activity. Taking the product of activity of sending and receiving neurons roughly tracks their correlations during each phase. Taking the difference of this correlation in each phase forces the network to unlearn its natural response and learn the desired target activity. In a spiking network, activity in these phases can be represented by low pass filters of spike trains; however, non-spiking activity can be assumed to approximate a rate-coding of spiking activity that fits some non-linear activation function. Unlike backpropagation, however, the network architecture \emph{requires} bidirectional synaptic connectivity (as shown between layer 1 and 2 of Fig.~\ref{fig:LocalLearning}~(a)) such that information propagates in both directions. Because each neuron is asynchronous, recurrence does not increase computational complexity as it does on traditional computer architectures. Additionally, the locality of learning and agnosticism to the neuron nonlinearity is advantageous for spiking neuromorphic hardware.

Following the two-layer network structure depicted in Fig.~\ref{fig:LocalLearning}~(a), we simulated an implementation of CHL on an ideal MZI-mesh neural network. A set of 40 input-output pairs were generated from randomly-distributed, uniform-magnitude, four-dimensional vectors. Each layer was simulated with four rate-coded neurons with a sigmoidal activation function; as such, each MZI mesh was simulated as a $4\times 4$ rectangular mesh. As in Fig.~\ref{fig:mzi}~(b), it is assumed that each MZI unit of each mesh contains four PDs for input and output monitoring. For simplicity, it is assumed that each neuron injects light to the mesh on a separate wavelength, and that the PD capacitance is large enough to reject the cross-term products between signals. Thus, the PD is assumed to linearly sum the power received from each wavelength. Because CHL assumes real-valued activation, phase shifter $\phi$ is neglected such that phase of each signal can be ignored. Following these assumptions, each MZI unit can be treated as a $2\times2$ sub-network that applies the following transformation to signal amplitude at each arm:
\begin{equation} \label{eq:Mat}
    \boldsymbol{W} = \left[\begin{matrix}
                        w_{11} & w_{12} \\
                        w_{21} & w_{22} \\
                     \end{matrix}\right]
                   = \left[\begin{matrix}
                        \text{sin}(\theta/2) & \text{cos}(\theta/2) \\
                        \text{cos}(\theta/2) & -\text{sin}(\theta/2) \\
                     \end{matrix}\right]
\end{equation}

Given that CHL is agnostic to the neural nonlinearity, Eq.~\ref{eq:CHL} can be directly applied to the photodetector outputs as long as they are measured correctly at the plus and minus phase. However, as seen in Eq.~\ref{eq:Mat}, the MZI mesh is not able to implement any arbitrary matrix. To resolve this, we can calculate derivatives that relate how a change in $\theta$ affects each individual weight. Next, we average the contribution from each $\Delta w_{ij}$ to estimate the best overall change:
\begin{equation}\label{eq:dTheta}
    \Delta\theta = \frac{1}{4}\sum_{i,j}\left[\left(\frac{dw_{ij}}{d\theta}\right)^{-1}\Delta w_{ij}\right]
\end{equation}

\begin{figure*}
    \centering
    \includegraphics[width=\linewidth]{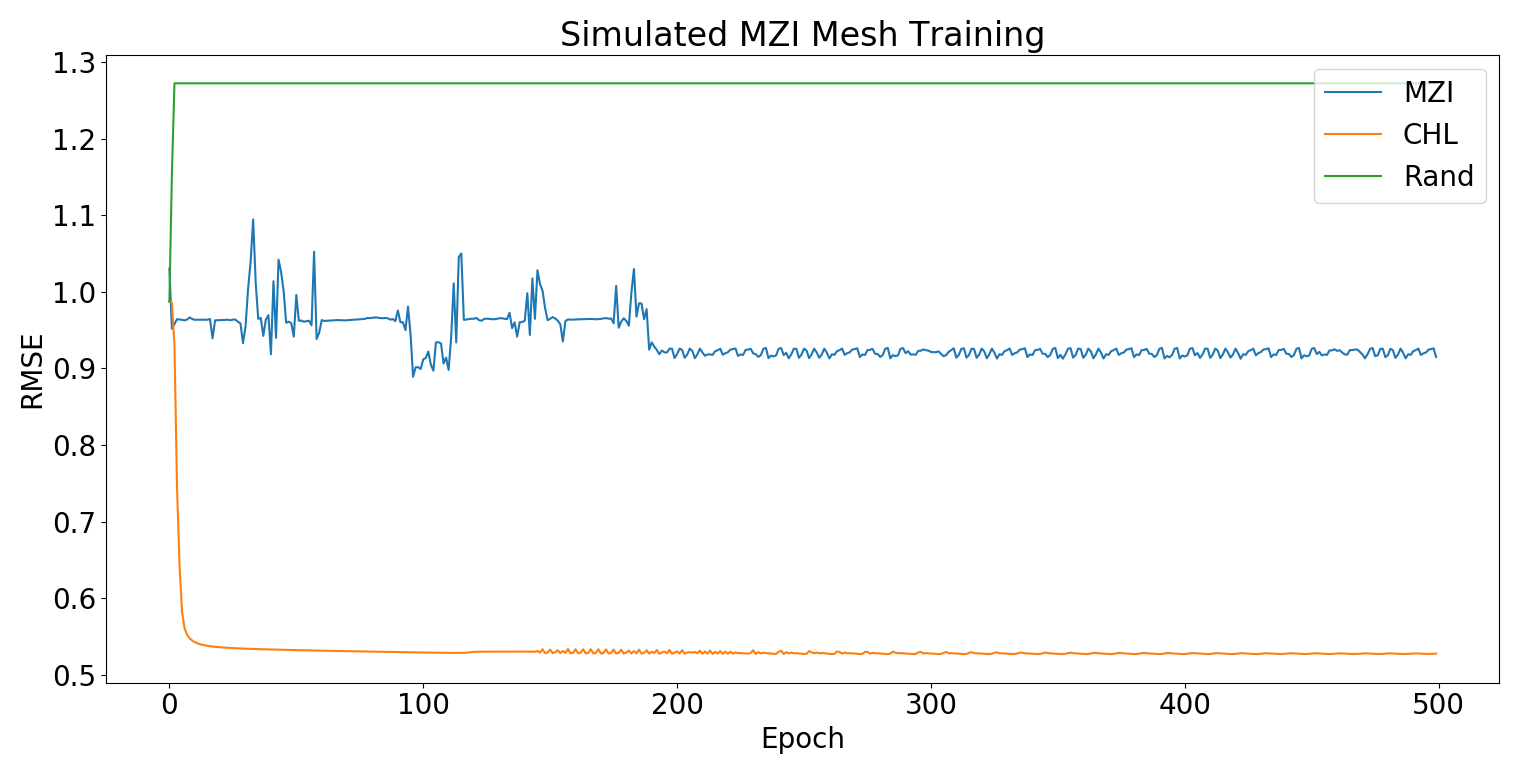}
    \caption{Root-mean-squared-error (RMSE) of the MZI-mesh network (blue) compared to an ideal implementation of CHL (orange) and another implementation with completely random learning signals (green).}
    \label{fig:CHL-MZI}
\end{figure*}

Note, we use $\left(dw_{ij}/d\theta\right)^{-1}$ because it is simpler to calculate than $d\theta/dw_{ij}$. Assuming that the plus and minus activity of each detector is recorded locally, this rule can be applied to every MZI in each mesh \emph{all at once}. Fig. \ref{fig:CHL-MZI} shows the root-mean-squared-error (RMSE) over each epoch for the aforementioned two-layer $4\times4$ network, along with an ideal implementation (direct application of Eq.~\ref{eq:CHL}) and implementation with randomly selected $\Delta w_{ij}$. Learning is applied after each sample (not batched) with $500$ epochs of training and a learning rate, $\eta=0.1$. Each implementation is initialized to the same starting matrices.

Our MZI implementation of CHL showed an $11.21\%$ decrease in RMSE over the course of training which is indicative of learning. However, the ideal implementation showed a significantly larger decrease of RMSE at $46.53\%$. For comparison, the randomly varying network shows an \emph{increase} of $28.87\%$ in RMSE, giving more credibility to the idea that the MZI-CHL implementation is capable of learning---albeit at a much slower rate than the ideal implementation. It is clear from the stochastic nature of RMSE in Fig~\ref{fig:CHL-MZI} that this implementation is prone to local minimums and instability. Nonetheless, this simple simulation illustrates the ability of the CHL rule to train local synaptic weights without regard for the other connections in the synaptic mesh and provides proof of concept for its use in MZI meshes. Additionally, the MZI mesh is restricted to unitary matrices which preserve magnitude of the input vector (before neural nonlinearity). In contrast, the ideal implementation allows independent gain and attenuation of each weight in the synaptic network. More work is needed to determine strategies for mitigating these restrictions and characterizing learning with more bio-realistic neural nonlinearities.

\subsubsection{Predictive Error-Driven Learning}
In the biology, however, target signals can only come from the network's own activity in response to its observations; even in the case of instructed learning, a biological brain must interpret perceptual stimuli (i.e. auditory and visual) and transform them into intelligible target signals for training. More recent work by O'Reilly \textit{et al.} \cite{Oreilly2021DeepPulvinar} has shown that the human brain may generate its own target training signals through cortico-thalamo loops which constantly undergo phases of prediction and observation to reduce future errors in prediction.
O'Reilly \textit{et al.} postulate that the alpha cycle ($\approx 10$ Hz) in the human brain demarcates iterations of such predictive error-driven learning, where plus and minus phases are separated by a bursting skip connection between primary processing regions and prediction-carrying regions (shown as the dotted connection in Fig.~\ref{fig:LocalLearning}~(b)) that fire with a $25\%$ duty-cycle within the alpha rhythm; a simplified diagram of this neural network architecture can be seen in Fig.~\ref{fig:LocalLearning}~(b).
Over many iterations of such prediction and observation, abstract representations can be learned that are capable of transformation-invariant object-recognition\cite{Oreilly2021DeepPulvinar}. The learning rule in this model is more complicated than CHL to include additional biologically relevant terms, though the error-driven learning is captured sufficiently by the simpler rule.

Bursting is important to enforce the 25\% duty-cycle and thus generate activity differences between the plus and minus phases of the CHL rule. The skip connection between first-order processing and the prediction layers allows the representation of the latter to more accurately match the ground-truth observation in the plus phase. Thus, without an explicit target signal, the network learns to better predict future inputs. Because subsequent inputs are governed by causality and are constantly occurring, the network is also constantly learning to better understand its environment. This structure can even be repeated for higher-order processing layers to hierarchically form even deeper, more abstract predictions of the input space. Future work is needed to identify an optimal implementation of the CHL rule within the MZI mesh structure and subsequently employ this style of self-supervised learning.

\section{Perspectives and Future Directions} \label{Perspectives}
    \subsection{Our Future System and Benchmarking}
The nanophotonic-electronic spiking neuron is composed of three main components: a photodetector, a nonlinear electrical circuit, and a laser. The photodetector receives information from the synaptic network and converts the optical signal to an electrical signal. The electrical circuit is the core of the neuron and processes the inputs to generate output spike responses. The laser output regenerates signal power after each layer to supply synaptic fanout to subsequent layers. Our team will exploit attojoule photonics with quantum impedance conversion \cite{DeLeonardis2019BroadbandMaterial} and closely integrate with low-capacitance ($<1\,fF$) electronics for monolithic integration on a silicon-on-insulator (SOI) platform. Using photonic communication between each SNN layer reduces capacitive charge associated with the interconnect wires \cite{Miller2017AttojouleCommunications} in comparable electronic circuits. Additionally, the photonic platform can allow neurons to communicate with other neurons at high speeds ($\sim$ 10 GHz) independently of communication distance.

To calculate the projected energy consumption, we can examine the composition of each component in the attojoule nanophotonic-electronic spiking neuron design. The dynamic energy cost of the nonlinear electronic circuit and laser can be calculated by examining the transistor on-state currents and associated operation voltages and frequency. Meanwhile the parasitic energy cost can be calculated from the total capacitance and the leakage current. According to our previous work\cite{Lee2022PhotonicModel}, the electrical circuit current flow inside the maximum $10\,GHz$ spiking rate attojoule neuron is expected to be $31.27\,\mu A$ at $1.4\,V$ voltage supply when the neuron is in the ON state, while the leakage current is $10\,nA$ in the OFF state. The expected nanolaser energy consumption is ~4.4 fJ per spike for a fanout of $ \sim $ 80 \cite{Miller2017AttojouleCommunications}. The parasitic capacitance includes the load capacitance on the photodetector, membrane capacitor, and transistor gate capacitance. According to the IRDS2020 \cite{2020THE2020} and \cite{Shastri2018a}, we expect the load capacitance of the photodetector to be around 0.1fF, and the simulated membrane capacitor to be 0.5fF. By considering closely integrated nanoelectronics at 10 fJ/bit energy efficiency and a fanout of 10-100 following the concept outlined by \cite{Miller2017AttojouleCommunications}, the minimum dynamic input energy to generate a spike output is projected to be 200aJ/spike.

For input, the proposed attojoule neuron design will utilize a low-Q nanophotonic crystal photodector with a Ge/Si cavity. The photonic crystal creates a resonant cavity that increases the confinement of light and reduces the size of the absorption medium \cite{El-Batawy201613Modeling}\cite{Nozaki2013InGaAsHeterostructure}. This allows for an ultra-low capacitance ($ \sim $0.1fF) nano-cavity PD that can generate sufficiently large voltage without amplification when combined with a high-impedance load \cite{Miller1989}. In addition, minimizing the electrical wiring between PDs and the nonlinear electronic circuit also reduces power consumption \cite{Miller2017AttojouleCommunications}. Similarly for spiking output, a hybrid InAs/AlGaAs quantum-dot nanolaser with photonic crystal cavity can be employed.


Aside from neuron design, the scalability on interconnect is also a critical design challenge. MZI meshes show nearly lossless  multiplication that is particularly suitable for large-scale low-power neuromorphic computing. However, the number of tunable elements, $N\cdot(N-1)$ in an $N\times N$ MZI mesh, grows polynomially with the number of neurons in the layer. As such, a control circuit must be designed that scales with a minimal additional computational complexity.
    
    \subsection{Footprint Efficiency}
    In the previous sections, we introduced and experimentally demonstrated bio-inspired on-chip training methods which improve the scalability of the SiPh MZI meshes for synaptic networks. We also simulated optoelectronic spiking neurons in GF 45SPCLO electronic-photonic hybrid platform and envisioned a scalable attojoule nanophotonic-electronic neuron design. However, one handicap of the proposed photonic neuromorphic system remains unaddressed, footprint efficiency. From our experience with commercial SiPh foundries, a 16$\times$16 MZI Meshes occupies a 12.5$mm^2$ chip area. Similarly, Lightmatter introduced their 64$\times$64 SiPh AI accelerator occupying a 150$mm^2$ chip area \cite{Ramey2020Silicon32} which incorporates billions of transistors. We propose two solutions, \textit{ Tensorized Photonic Neural Networks} (TPNN) and \textit{3D Electronic-Photonic Integrated Circuits} (3D EPICs) to improve footprint efficiency and enable deep and wide photonic neuromorphic systems. 
        \subsubsection{TPNN}
        There are three main methods to avoid over-parameterized neural networks and relieve hardware requirements such as \textit{weight pruning}, \textit{quantization}, and \textit{model compression} \cite{Han2015LearningNetworks}. Because photonic NNs are analog computers, available bit precision is already limited. Unlike electronics, a photonic system can easily offer all-to-all connectivity through wavelength and space-division multiplexing. Therefore, the benefits of weight pruning and quantization approaches are not significant. In contrast, model compression can result in fewer hardware resources and smaller footprints. We proposed and simulated an algorithm-hardware co-design approach: photonic tensorized neural networks \cite{Xiao2021ScalableNetworks}.
        Tensor-Train (TT) decomposition is a multi-dimensional array processing technique to represent large matrices in a low-rank approximation \cite{Oseledets2011Tensor-trainDecomposition}. Although low-rank approximation may cause decreased performance in NNs, one could train NN models in TT-decomposed format so that performance degradation is minimized \cite{NovikovTensorizingNetworks}. For some ML problems, low-rank approximation also serves as a regularization term and improves performance \cite{Hawkins2021BayesianSelection}. Moreover, in the simulations \cite{BerkayOn2021AnalysisNetworks}, we observed that TT-decomposed MZI meshes are more resilient to noise and hardware imprecision. Our simulations and benchmarks demonstrated that TPNN could improve the footprint-energy-efficiency product by $4$ orders of magnitude by using $79\times$ fewer 2$\times$2 MZI units without decreasing accuracy below 95\% in image classification tasks \cite{Xiao2021Large-scalePlatform}. Future work will realize a SiPh end-to-end TPNN system and provide benchmarks for footprint-energy efficiency and performance.

        \subsubsection{3D EPIC}
        3D electronic ICs (EIC) promise low energy consumption, low noise, and high density because of shorter electrical wires \cite{Knickerbocker2008Three-dimensionalIntegration}. The main enabling technology for 3D EICs is through-silicon vias (TSV). Although thermal relief and yield are the challenges, 3D integrated high bandwidth memories show clear advantages compared to 2D EICs. Similarly, 3D electronic-photonic ICs (EPICs) can achieve high density, low loss, and high bandwidth performance. Multi-layer silicon photonic devices are already available in commercial foundries. However, they rely on evanescent vertical couplers, which require relatively long taper lengths ($\sim 100\,\mu m$) and small layer distance ($\sim 1\,\mu m$) \cite{Sacher2017Tri-layerTransitions, Shang2015Low-lossCircuits}. As an alternative, our previous work demonstrates through silicon optical vias (TSOV) \cite{Zhang2020ScalableApplications, Zhang2018High-DensityCircuits} for 3D EPICS using 45degree reflectors and silicon vias\cite{Zhang2018High-DensityCircuits}. Ultrafast laser inscription also allows for freeform shaping of waveguides useful for routing in three dimensions. This technique has already been demonstrated for orbital-angular momentum multiplexing/demultiplexing and optical beam steering applications\cite{BenYoo2016HeterogeneousMicrosystems}. 3D EPICs provide devices to be stacked vertically allowing for greater neuron density per area and thus the design of deeper and wider photonic neural networks.
        
    \subsection{Applications for SNNs}
        In relation to AI and machine learning, SNNs provide several advantages over modern computing paradigms for tasks which mimic the conditions in which they naturally evolved. Because SNNs process data over time in a continuous manner, they are well-suited to applications situated in real-time environments with single inference and learning instances presented at a time (such as event-based signal processing \cite{Blouw2020Event-DrivenSystems}). In addition, the spread of information over time allows multiple forms of memory at different time-scales similar to the human distinction between working \cite{Giulioni2012RobustVLSI}, short-term \cite{Rao2022AHardware}, and long-term memories. Neuromorphic sensing and robotics are a common direction of applications of SNNs; for example, an adaptive robotic arm controller can provide reliable motor control as actuators wear down \cite{DeWolf2016AControl}. More speculatively, future devices might exploit these properties in the context of live audio and natural language processing for voice assistants, live-captioning services, or audio separation; similarly SNNs can be used for live video and lidar processing in autonomous vehicles or surveillance systems. SNNs are not ideal for batched computation---in which multiple training samples are computed in parallel and averaged for parallelism in training---however, data centers may still make use of the increased computational parallelism in tasks like the nearest-neighbor search which can be performs in constant time, $O(1)$, on neuromorphic chips like Loihi \cite{Frady2020NeuromorphicSprings}.
        
        A major challenge of many modern DNN and reinforcement learning (RL) agents is the development of abstract, transformation-invariant representations of objects relevant to the task. In classification tasks, a neural network must transform its input space into a representation which most clearly separates each labeled class. Similarly, RL agents must be able to process their input space into a representation that best accentuates the value of potential actions. Predictive error-driven learning, modeled after the work of O'Reilly \cite{Oreilly2021DeepPulvinar}, has the potential to autonomously build deep hierarchies of abstraction for a given input space. For example, a learning agent could implicitly learn physical properties of the world such as gravity, buoyancy, and contact forces simply by observing its environment. In combination with complimentary learning systems for memory \cite{OReilly2014ComplementarySystems} and RL models based on the basal ganglia \cite{Rasmussen2017ALearning}, a neuromorphic learning agent may be capable of replicating simple navigation and foraging behaviors which require the flexible application of knowledge and memory. Such a model could provide key insights for the development of self-motivated learning agents that exploit hierarchical representations to solve reinforced tasks. Developing dedicated spiking neuromorphic hardware and taking advantage of the energy-efficient and scalable photonic devices will allow the development of larger models and new computational paradigms. These developments can be applied in dynamic, noisy environments that are not well-handled by today's machine learning efforts.

\section{Conclusion}
We have discussed the advantages of dedicated SNN hardware and highlighted the benefits of nanophotonic-electronic design within this computational paradigm. Additionally, we argued that co-integration of photonic and electronic devices combines the high-bandwidth, low-power communication protocols of photonics with the well-established and flexible CMOS circuitry. Towards the construction of a photonic SNN computing architecture, we demonstrated an Izhikevich-inspired optoelectronic neuron design, implemented RPB on an MZI mesh, and simulated CHL on a rate-coded, MZI-mesh neural network. In addition, we proposed the construction of a powerful self-learning SNN computing architecture built from these technologies and based on predictive error-driven learning models of the human brain. Subsequently, we have discussed technologies for improving the scalability of neuron and network density through tensorization of large neural networks and 3D electronic-photonic integration. Finally, we discussed perspectives on the suitable applications of photonic SNNs and emphasized applications of interest for our own efforts. 

Future work is needed to establish the optimal design for brain-inspired spiking networks. Modern ANNs have oversimplified neural nonlinearities due to the limitations of the von Neumann computing architecture. Meanwhile, the heterogeneity of neural behaviors in different regions of the human brain provide various methods of encoding information. As such, a deeper exploration of these encodings is warranted to fully leverage the computing power of SNNs. Furthermore, modern learning algorithms are designed for sequential processing that is not ideal for SNNs hardware. As such, considerable work is necessary to determine the most efficient on-chip implementation of local learning rules like CHL. Nonetheless, the design challenges are well worth the effort to provide alternative routes for continued advances in computation and signal processing in the face of slowing progress of transistor scaling. Our continued work will focus on the characterization and design of nanophotonic-electronic spiking neurons and their incorporation within scalable, MZI-based neural networks capable of on-chip local learning.


%

\section*{Acknowledgment}

This work was funded in part by the Air Force Office of Scientific Research grant FA9550-181-1-0186.

This research is based upon work supported in part by the Office of the Director of
National Intelligence (ODNI), Intelligence Advanced Research Projects Activity (IARPA), via
[2021-21090200004]. The views and conclusions contained herein are those of the authors
and should not be interpreted as necessarily representing the official policies, either
expressed or implied, of ODNI, IARPA, or the U.S. Government. The U.S. Government is
authorized to reproduce and distribute reprints for governmental purposes notwithstanding
any copyright annotation therein.

The authors would like to thank GLOBALFOUNDRIES for providing silicon fabrication through the 90WG university program and for their technical assistance in 45SPCLO MPW runs.

\appendices
\section{RBP Algorithm}\label{RBP_Pseudocode}
\begin{algorithm}
	\caption{Random Backprop on SiPh MZI Mesh}
	\label{alg:the_alg}
	\begin{algorithmic}[1]
	    \State Initialize resistor values $\mathbf{R}$, accuracy limit $L$, total number of samples $N$, MZI voltages $\mathbf{v}^{-1}_{MZI}$, error $\mathbf{e}^{-1}=\infty$, coarse and fine step sizes $\mu_{c}, \mu_{f}$, start with coarse search $\mu\leftarrow\mu_{c}$, , random backprop weights $\mathbf{B}\sim[-\mu,\mu]$
		\For {Every epoch}
		\For {$k=0$ through $N$}
		\State Find input generator voltages $\mathbf{v}^k_{in}$ for $\mathbf{x}^k$ in $LUT$
		\State Read input generator's PDs to verify $\mathbf{x}^k$
		\State Read output PDs $\mathbf{v}_{out}$
		\State Calculate photocurrent $\mathbf{i}_{out}=(v_{dd}-\mathbf{v}_{out})/\mathbf{R}$
		\State Normalize $\mathbf{i}_{out}$ to calculate $\mathbf{\hat{y}}^k$
		\State Calculate error $\mathbf{e}^{k}=|\mathbf{\hat{y}}^k-\mathbf{y}^k|^2$
		\If {$\mathbf{e}^{k}>\mathbf{e}^{k-1}$}
		\State Draw a new $\mathbf{B}\sim[-\mu,\mu]$
		\EndIf
		\State Update $\mathbf{v}^{k}_{MZI}\leftarrow\mathbf{v}^{k-1}_{MZI}+\mathbf{B}\mathbf{e}^{k}$
		\EndFor
		\State Calculate interference accuracy $a$
		\For {Every sample $\mathbf{x}^k$}
        \State Find input generator voltages $\mathbf{v}^k_{in}$ for $\mathbf{x}^k$ in $LUT$
		\State Read input generator's PDs to verify $\mathbf{x}^k$
		\State Read output PDs $\mathbf{v}_{out}$
		\State Calculate photocurrent $\mathbf{i}_{out}=(v_{dd}-\mathbf{v}_{out})/\mathbf{R}$
		\State Decide class label $\hat{l}^k=\underset{n}{\arg\max}\ {i}_{out}[n]$
		\EndFor
		\State $a=sum(\mathbf{\hat{l}} == \mathbf{l})$
		\If {$a \geq L$}
		\State Switch to fine search $\mu\leftarrow\mu_{f}$
		\EndIf
	\EndFor
	\end{algorithmic} 
\end{algorithm}

\ifCLASSOPTIONcaptionsoff
  \newpage
\fi





%

%


\begin{IEEEbiographynophoto}{Yun-Jhu Lee}
received the B.S. in Life Science from the National Taiwan University, Taiwan. He is currently working towards the Ph.D degree in Electrical and Computer Engineering at the University of California, Davis. Research interests include neuromorphic computing, integrated photonics, MEMS, and control system.
\end{IEEEbiographynophoto}

\begin{IEEEbiographynophoto}{Mehmet Berkay On}
received the B.S. in Electrical and Electronics Engineering from the Bilkent University, Ankara, Turkey in 2018. He is currently working towards the Ph.D degree in Electrical and Computer Engineering at the University of California, Davis. Research interests are enerrgy-efficient photonic neuromorphic systems, RF-photonic signal processing, fiber-optic communication, and compressive sensing.
\end{IEEEbiographynophoto}


\begin{IEEEbiographynophoto}{Luis El Srouji}
received the B.S. in Applied Physics with an emphasis in Physical Electronics from the University of California, Davis in 2020. He is currently working towards the Ph.D degree in Electrical Engineering at the University of California, Davis. Research interests include the design of bio-physically accurate analog neuron circuits, development of learning algorithms for optoelectronic spiking neural networks, and fabrication of on-chip laser sources.
\end{IEEEbiographynophoto}

\begin{IEEEbiographynophoto}{Li Zhang}
Li Zhang received the B.S. degree in electronics and information technology and instrumentation from Zhejiang University, Hangzhou, China, in 2016. He is currently pursuing the Ph.D. degree in electrical engineering with the University of California at Davis, Davis, CA, USA. His research interests include ultra-wideband transceiver, trans-impedance amplifier and optical driver.
\end{IEEEbiographynophoto}

\begin{IEEEbiographynophoto}{S. J. Ben Yoo}
(Fellow, IEEE and Fellow, Optica) received the B.S. degree in electrical engineering with distinction, the M.S. degree in electrical engineering, and the Ph.D. degree in electrical engineering with a minor in physics, from Stanford University, Stanford, CA, USA, in 1984, 1986, and 1991, respectively. He is currently a Distinguished Professor of electrical engineering with UC Davis, Davis, CA, USA. His research with UC Davis includes 2D/3D photonic integration for fu- ture computing, communication, imaging, and navigation systems, micro/nano systems integration, and the future Internet. Prior to joining UC Davis in 1999, he was a Senior Research Scientist with Bellcore, leading technical efforts in integrated photonics, optical networking, and systems integration. His research activities with Bellcore included the next-generation internet, reconfigurable multiwavelength optical networks (MONET), wavelength interchanging cross connects, wavelength converters, vertical-cavity lasers, and high-speed modu- lators. He led the MONET testbed experimentation efforts, and participated in ATD/MONET systems integration and a number of standardization activities. Prior to joining Bellcore in 1991, he conducted research on nonlinear optical processes in quantum wells, a four-wave-mixing study of relaxation mechanisms in dye molecules, and ultrafast diffusion-driven photodetectors with Stanford University. He is a fellow of OSA, NIAC, and was the recipient of the DARPA Award for Sustained Excellence (1997), the Bellcore CEO Award (1998), the Mid-Career Research Faculty Award (2004 UC Davis), and the Senior Research Faculty Award (2011 UC Davis).
\end{IEEEbiographynophoto}




\end{document}